\documentclass[journal]{IEEEtran}
\usepackage{times}
\usepackage{epsfig}
\usepackage{graphicx}
\usepackage{amsmath}
\usepackage{amssymb}

\usepackage{booktabs}
\usepackage{soul}
\usepackage{algorithm}
\usepackage{algorithmic}
\usepackage{xcolor}
\usepackage{bm}
\usepackage{multirow}
\usepackage{url}

\newcommand{\tabincell}[2]{\begin{tabular}{@{}#1@{}}#2\end{tabular}}
\newcommand{\para}[1]{\noindent\textbf{#1}\ \ }
\newcommand{\dif}[1]{{#1}}
\newcommand{\rebuttal}[1]{{#1}}
\newcommand{\thirdrebuttal}[1]{{#1}}

\begin{document}
%
\title{DAIS: Automatic Channel Pruning via Differentiable Annealing Indicator Search}
%
%
%

\author{

    Yushuo Guan,
    Ning Liu,~\IEEEmembership{Member,~IEEE}, 
    Pengyu Zhao, 
    Zhengping Che,~\IEEEmembership{Member,~IEEE},\\
    Kaigui Bian,~\IEEEmembership{Senior Member,~IEEE}, 
    Yanzhi Wang,~\IEEEmembership{Member,~IEEE},
    and Jian Tang,~\IEEEmembership{Fellow,~IEEE}
\thanks{Yushuo Guan and Pengyu Zhao are with the School of Computer Science, Peking University, Beijing, China 100871.
E-mail: \textit{\{david.guan, pengyuzhao\}@pku.edu.cn}}
\thanks{Ning Liu, Zhengping Che, and Jian Tang are with Midea Group, Beijing, China 100102.
E-mail: \textit{\{liuning22, chezp, tangjian22\}@midea.com}}
\thanks{Kaigui Bian is with the School of Computer Science, National Engineering Laboratory for Big Data Analysis and Applications, Peking University, Beijing, China 100871.
E-mail: \textit{bkg@pku.edu.cn}}
\thanks{Yanzhi Wang is with the Department of Electrical \& Computer Engineering, Northeastern University, Boston, MA, USA 02115.
E-mail: \textit{yanz.wang@northeastern.edu}}
\thanks{Corresponding author: Kaigui Bian.}
}

\maketitle

\begin{abstract}
    The convolutional neural network has achieved great success in fulfilling computer vision tasks despite large computation overhead against efficient deployment.
    Channel pruning is usually applied to reduce the model redundancy while preserving the network structure, such that the pruned network can be easily deployed in practice.
    \rebuttal{However, existing channel pruning methods require hand-crafted rules, which can result in a degraded model performance with respect to the tremendous potential pruning space given large neural networks.}
    In this paper, we introduce Differentiable Annealing Indicator Search (DAIS) that leverages the strength of neural architecture search in the channel pruning and automatically searches for the effective pruned model with given constraints on computation overhead.
    Specifically, DAIS relaxes the binarized channel indicators to be continuous and then jointly learns both indicators and model parameters via bi-level optimization.
    To bridge the non-negligible discrepancy between the continuous model and the target binarized model, DAIS proposes an annealing-based procedure to steer the indicator convergence towards binarized states.
    Moreover, DAIS designs various regularizations based on \emph{a priori} structural knowledge to control the pruning sparsity and to improve model performance.
    Experimental results show that DAIS outperforms state-of-the-art pruning methods on \hbox{CIFAR-10}, CIFAR-100, and ImageNet. 
\end{abstract}


\section{Introduction}
In recent years, the community has achieved great success on various computer vision tasks~\cite{goodfellow2014generative,deepface,hourglass,he2017mask} by designing deeper and wider convolutional neural networks~(CNNs). Despite the superior performance of these networks, the computation cost is burdensome. Hence, it is difficult to directly deploy such ``expensive'' networks over computation-limited \rebuttal{applications} such as robotics, self-driving vehicles, and most of the mobile devices. 
A straightforward solution to tackle the problem is \textit{channel pruning}. \dif{Channel pruning methods target at compressing the networks \thirdrebuttal{in terms of} parameters, computation cost, and inference latency, while maintaining the performance of the unpruned networks~\cite{abs-2101-09671,liu2017learning, NIPS2019_8364}. Channel pruning methods could reduce the computation cost of the network by eliminating redundant channels \rebuttal{in CNNs}, \thirdrebuttal{which require} no extra hardware support and is perfectly compatible with representative deep learning acceleration frameworks, such as TVM~\cite{chen2018tvm}, TFLite~\cite{TensorFlow-Lite}, and Alibaba MNN~\cite{Ali-MNN}.}

Earlier channel pruning methods leverage hand-crafted criteria, such as lasso regression~\cite{liu2017learning} and geometric median-based criterion~\cite{he2019filter}, to filter out unimportant channels. Recent research on pruning~\cite{liu2018rethinking} implies that the pruning process is equivalent to searching a compact network structure, suggesting the possibility of automatic channel pruning.
\dif{
Since the pruning search space increases exponentially with original networks being deeper and wider, a potential approach to address this problem is through the idea of automatic channel pruning~\cite{he2018amc,NIPS2019_8364}. These methods \rebuttal{search} for the appropriate pruned models automatically, but the search efficiency might be low~\cite{he2018amc} and there might be architecture discrepancies between the search process and the final pruned model~\cite{NIPS2019_8364}.}

To address these problems in the automatic pruning approaches, we propose DAIS, a Differentiable Annealing Indicator Search method for channel pruning, taking advantage of differentiable neural architecture search~\cite{liu2019darts}. In general, DAIS first searches for a pruned model in an automatic manner with computational constraints and then fine-tunes the derived model.
Specifically, DAIS introduces binarized channel indicators and utilizes continuous auxiliary parameters to relax the indicators for optimization.
Then it jointly learns model parameters and auxiliary parameters via bi-level optimization, to simultaneously obtain the accurate model and identify the importance of each channel.
An annealing-relaxed function is incorporated into the channel indicators to mitigate the discrepancy between the derived pruned model and the pretrained supernet for the search procedure.
The indicators are initialized to be gently continuous at high temperatures, and gradually converge to binarized states as the training proceeds and temperature anneals. 
Furthermore, to provide the structural constraints on the pruned model, we design dedicated regularizers in DAIS: 1) A continuous FLOPs estimator regularizer that controls the computational cost of the pruned model, and 2) a symmetry regularizer which optimizes the gradient propagation on the pruned ultra-deep neural networks with residual connections. \rebuttal{Additionally, DAIS is more efficient, since it is a one-shot solution that does not require multi-round pruning compared with most existing methods~\cite{liu2017learning,luo2017thinet,you2019gate}.}
Experimental results show that DAIS outperforms state-of-the-art pruning methods on representative datasets such as CIFAR-10~\cite{krizhevsky2009learning}, CIFAR-100~\cite{krizhevsky2009learning}, and ImageNet~\cite{deng2009imagenet}, \thirdrebuttal{as well as having the great transfer ability to other vision tasks such as semantic segmentation and scene text recognition}.

The main contributions of this paper are as follows. 1)~We propose DAIS, a differentiable annealing indicator search framework for channel pruning, which leverages the gradient-based bi-level optimization to search for appropriate pruned models with sparsity requirements. 2) We design an annealing-relaxed channel indicator function for the differentiable search process. \rebuttal{This function is initialized to be continuous to relax the search process to be differentiable and then gradually converges into binarization to produce the pruned model.} 3)~We design different regularizers to constrain the computational cost of the pruned model. 4) DAIS achieves state-of-the-art performance on different datasets and models, and extensive experiments and ablation studies demonstrate its effectiveness.

\dif{
\section{Related Works}

\para{Channel pruning.}
The recent advances of network pruning mainly fall into two categories: unstructured pruning and channel pruning. Unstructured pruning~\cite{han2015learning,guo2016dynamic} removes weights at arbitrary locations to reduce the storage and computation. However, the resulted sparse matrix and indexing scheme derived by unstructured pruning methods requires special sparse matrix operation libraries
and/or hardware, which limit the practical acceleration in general CNN acceleration frameworks. \rebuttal{To avoid these limitations, recent works~\cite{luo2017thinet,he2017channel,yu2018nisp,zhuang2018discrimination,he2019filter,liuautocompress} focus on channel pruning, which filter out redundant channels of the convolutional networks.} The pruned model would maintain the network structure and take full advantage of Basic Linear Algebra Subprograms (BLAS) operations~\rebuttal{\cite{blackford2002updated}}. Thus it can be perfectly supported by the prevalent CNN acceleration frameworks such as TVM~\cite{chen2018tvm}, TensorFlow-Lite (TFLite)~\cite{TensorFlow-Lite}, and Alibaba Mobile Neural Network (MNN)~\cite{Ali-MNN}.

Early works~\cite{wen2016learning,li2016pruning,he2018soft} heuristically evaluate the importance by the magnitude of the weights. Wen \textit{et al.}~\cite{wen2016learning} adopt group lasso to force groups of weights to be smaller and filter out the channels with zero weights. PFEC~\cite{li2016pruning} identifies the importance of each channel by L1 norm, and prunes the channels which are identified to \thirdrebuttal{have small impact} on the network performance. SFP~\cite{he2018soft} first selects the filters based on L2 norm and then prunes them softly. All these works use \thirdrebuttal{simple criteria like L1 and L2 norm} to determine the significance of channels and filter out channels with smaller norms.

Recent methods~\cite{he2019filter,li2019compressing,you2019gate,He_2020_CVPR,Lin_2020_CVPR,LinJZZW020,icml2020_1485} design various criteria or utilize additional optimization tools for channel pruning. FPGM~\cite{he2019filter} filters out redundant channels with a criterion of the geometric median. It proves that pruning the channels near the geometric median has a substantial impact on the network accuracy. CNN-FCF~\cite{li2019compressing} learns binary scalars associated with filters to determine the target filters to prune. The binary scalars are not differentiable, and additional optimization tools (such as ADMM) are needed for addressing the binary constraints. GBN~\cite{you2019gate} introduces the gate decorator into the network and estimates the performance of candidate pruned models based on Taylor expansion. LFPC~\cite{He_2020_CVPR} learns different criteria for each convolutional layer, considering that different layers have various distributions. HRank~\cite{Lin_2020_CVPR} discovers that the low-rank feature maps contain less information, therefore it prunes filters with low-rank feature maps. ABCPruner~\cite{LinJZZW020} leverages the artificial bee colony~\thirdrebuttal{(ABC) algorithm} to search for appropriate pruned models. \rebuttal{SCP~\cite{icml2020_1485} jointly considers the effect of batch normalization (BN)~\cite{ioffe2015batch} and ReLU activation, and filters out the channels where the BN and ReLU operations are likely to deactivate each feature map. }

Besides, some channel pruning approaches~\cite{dong2017more,huang2018data,lin2019towards} target at the efficiency of the pruning process. Dong \textit{et al.}~\cite{dong2017more} propose the low-cost collaborative layer (LCCL) to speed up the inference of CNNs. SSS~\cite{huang2018data} prunes the networks in one training pass with the help of a modified stochastic Accelerated Proximal Gradient (APG) method. Lin \textit{et al.}~\cite{lin2019towards} optimizes the channel pruning by generative adversarial learning (GAL), which replaces the traditional multi-stage pruning into the end-to-end optimization.

\para{Neural architecture search (NAS).} 
Recently, there has been a growing interest in neural architecture search (NAS), which automatically designs neural network architecture with no human effort. The neural architecture search approaches could be classified according to the \textit{search space} and \textit{search strategy}. 

The search space of NAS is the network topology, and could be categorized into the macro search space and micro search space. The methods~\cite{zoph2017neural,cai2018proxylessnas} search for the entire CNN architecture in the macro search space, including kernel sizes, skip connections and number of filters. Since the cost of the macro space search is too large, more methods focus on the micro space search, where they only search for a certain module structure in the network~\cite{liu2019darts,zoph2018learning}. The derived module will be stacked to form a complete network. Early works leverage reinforcement learning~\cite{zoph2017neural,baker2017designing,zoph2018learning,tan2019mnasnet} or the evolutionary algorithm~\cite{real2017large,real2019regularized} as the search strategy. These works mainly sample a large number of networks from search space, then train them from scratch to obtain a supervision signal, and optimize the sampling agent with different search strategies. \rebuttal{PGNAS~\cite{ZhouSLZZ20} models the NAS task from Bayesian perspective and needs a posterior-guided sampling process. It reduces the computational complexity compared with earlier approaches~\cite{zoph2017neural,zoph2018learning}, but still needs 11 GPU days to search for appropriate networks.} Recent attempts~\cite{bender2018understanding,pham2018efficient,chen2019detnas,li2019random,wu2019fbnet,xie2019snas} introduce weight-sharing paradigm in NAS to boost search efficiency, where all candidate sub-networks share the weights in a single one-shot model that contains every possible architecture in the search space.
Among weight-sharing methods, DARTS \cite{liu2019darts} has attracted much attention.
It relaxes the search space to be continuous with architecture parameters and then efficiently optimizes model parameters and architecture parameters together via gradient descent. 

\para{Automatic channel pruning.}
The channel pruning could be thought of searching for appropriate pruned models as suggested by \thirdrebuttal{Liu \textit{et al.}}~\cite{liu2018rethinking}, which implies the possibility of automatic channel pruning. The automatic channel pruning~\cite{he2018amc,NIPS2019_8364} leverages the idea of NAS, which spends less search cost compared with the complete NAS.
NetSlimming~\cite{liu2017learning} automatically selects filters by associating scaling factors from BN layers and prunes filters with smaller scaling factors. AMC~\cite{he2018amc} leverages deep reinforcement learning to determine the pruning rate of each layer. It uses a sampling, estimating, and learning process, which is time-consuming, especially for deep networks. Besides, AMC only prunes the middle channels of the blocks with shortcut (like ResNet), which limits the optimal upper bound of the pruning ratio. TAS~\cite{NIPS2019_8364} directly searches the width and depth of a network with a novel transformable architecture search procedure, but there are architecture discrepancies between the supernet in the search procedure and the pruned model, which 
may result in performance degradation in terms of pruning ratio and accuracy.

\begin{figure*}[t]
\centering
\includegraphics[width=0.99\linewidth]{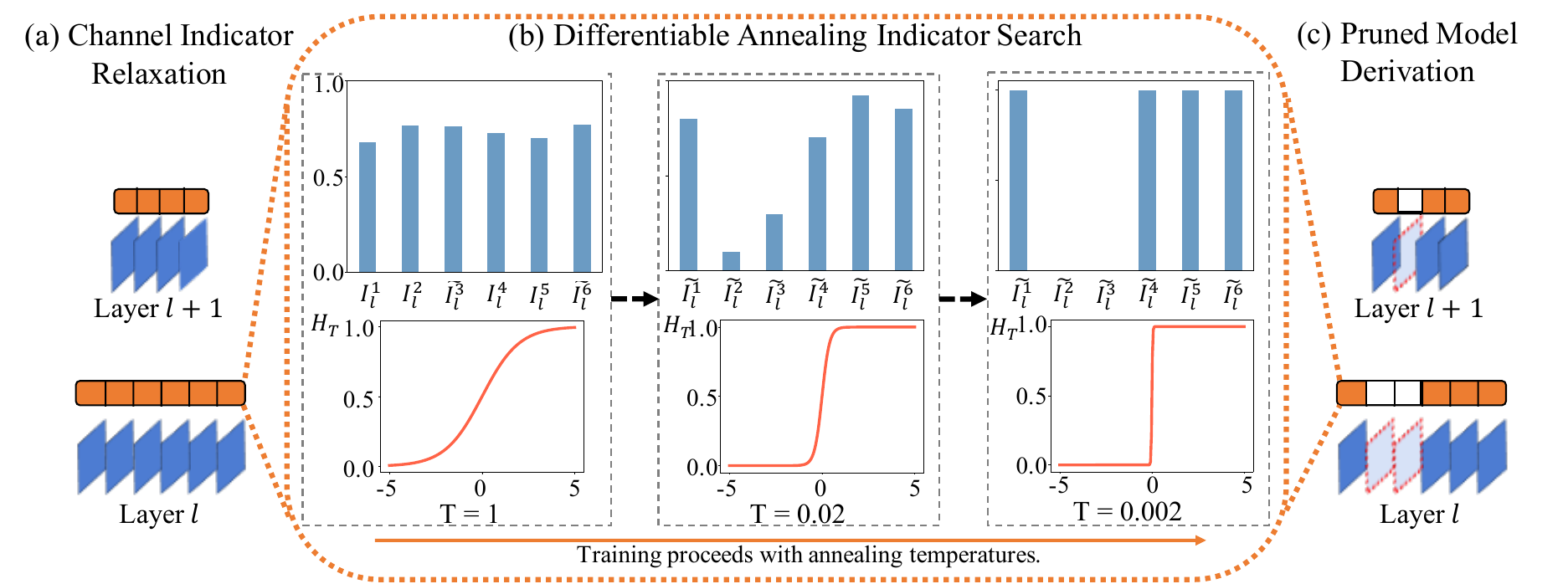}
\caption{Overview of the Differentiable Annealing Indicator Search (DAIS) framework. (a) Channel indicator relaxation. We build annealing-relaxed channel indicators for each convolutional layer. 
(b) Differentiable annealing indicator search. As the training proceeds, the annealing-relaxed indicator $\widetilde{I_l^i}$ will converge to binarization with annealing temperatures, as illustrated in the histograms. $H_{T}(*)$ is an annealing function, which approximates the binarized state when the temperature anneals. 
(c) Pruned model derivation. After the differentiable annealing indicator search, the pruned model will be derived by filtering out the output channels with $\widetilde{I_l^i}=0$ from the original model.}
\label{fig:overview}
\end{figure*}

\rebuttal{
\para{Comparisons to related works.}
DAIS leverages the automatic differentiable search procedure to replace heuristic rules in traditional pruning methods~\cite{luo2017thinet,he2017channel,yu2018nisp,zhuang2018discrimination,he2019filter,liuautocompress}. DAIS is one-shot, which is more efficient compared with multi-round pruning approaches~\cite{liu2017learning,luo2017thinet,you2019gate}. The heuristic annealing idea of DAIS could efficiently support the convergence of the bi-level pruning process. DAIS has different optimization targets compared with general NAS works like PGNAS~\cite{ZhouSLZZ20}. General NAS works focus on neural architectures with high accuracy, while DAIS targets at automatic channel pruning and searches for a small and fast pruned model for the original model. DAIS also breaks the limitations of automatic channel pruning methods like AMC~\cite{he2018amc} and TAS~\cite{NIPS2019_8364}. DAIS is built on top of the differentiable indicator search procedure for faster search efficiency, and leverages the annealing-relaxed channel indicator to fill the architecture discrepancies. DAIS also uses different regularizers to satisfy structural restrictions. 
}}

\section{Methodology}

In this section, we firstly formulate the channel pruning task and then introduce the Differentiable Annealing Indicator Search (DAIS) method for channel pruning. Specifically, we demonstrate the differentiable channel indicator search procedure of DAIS, propose the annealing-relaxed channel indicator, and introduce three regularizers for structural restrictions. The overview of DAIS is illustrated in Fig.~\ref{fig:overview}.

\subsection{Problem Definition}\label{sec:method:problem}
A general CNN model is built with a stack of convolutional layers. Suppose the model has $L$ layers, and we define $O_l$ as the output feature tensor of the $l$-th convolutional layer. 
The computation process of $O_l$ can be commonly depicted as:
\begin{equation}
    O_l = F_l(W_l, O_{l-1}),
\end{equation}
where $W_l \in \mathcal{R}^{k_l \times k_l \times c_{l-1} \times c_l}$ denotes the convolutional kernel of the $l$-th layer with input channel of $c_{l-1}$ and output channel of $c_l$, and
$F_l(\cdot)$ represents the convolution function.
The optimization target of model pruning is to derive the network architecture with the minimum number of convolutional filters from the original model while maintaining accuracy. 
The channel pruning further preserves the structural information of CNN and only reduces the number of channels, i.e., $c_l$ for each layer.

A popular channel pruning approach~\cite{liu2017learning} introduces channel indicators, which are binarized vectors generating sparsity for each convolutional layer, i.e.:
\begin{equation}
    O_l^{'} = F_l(W_l, O_{l-1}^{'}) \otimes I_l,
\end{equation}
where $I_l \in \{0,1\}^{c_l}$ denotes the indicator vector, $\otimes$ represents the tensor product, and $O_l^{'}$ denotes the masked (pruned) output tensor by $I_l$. 
With channel indicator, the optimization target of channel pruning can be represented by:
\begin{equation}
    \min_{\mathcal{W}, \mathcal{I}} \ \mathcal{L}(\mathcal{W}, \mathcal{I}) + \lambda \mathcal{R}(\mathcal{I}),
\end{equation}
where $\mathcal{W} = [W_1, W_2, ..., W_L]$ and $\mathcal{I} = [I_1, I_2, ... I_L]$ denote respectively the parameter and indicator sets, $\mathcal{L}(\cdot)$ denotes the loss function, and $\mathcal{R}(\cdot)$ denotes the regularizer that induces the structural restrictions. The details of regularizers will be discussed in Sec.~\ref{sec:method:sr}.

As the channel indicator is binarized and non-differentiable, it could not be learned directly by conventional gradient descent methods. 
Therefore, some channel pruning approaches~\cite{liu2017learning, you2019gate} associate the channel indicator with differentiable metrics, e.g., leverage the weights in \rebuttal{BN} as an indirect indicator, and zero out the channels with small BN weights~\cite{liu2017learning}. 
Instead of building up these indirect associations, we deal with the channel indicator from another perspective which is more effective and flexible, i.e., \emph{relax the channel indicators to be continuous, and then jointly learn the model parameters and relaxed channel indicators via a differentiable search procedure}.

\subsection{Differentiable Indicator Search}\label{sec:method:DAIS}

The differentiable search procedure is widely used in NAS~\cite{chu2019fair,liu2019darts,wu2019fbnet} but seldom explored towards solving the channel pruning problem. 
In this section, we incorporate the idea of differentiable search into channel pruning and introduce the differentiable indicator search.

To make the search space continuous, we relax each binarized channel indicator $I_l^i$, $i \in [1,c_l]$, into a relaxed channel indicator $\widetilde{I}_l^i$ parameterized by an auxiliary parameter $\alpha_{l}^i$ (discussed in Sec.~\ref{sec:method:aci}).
After the relaxation, the goal of the indicator search is to jointly learn model parameters $\mathcal{W}$ and auxiliary parameters $\alpha$. 
An intuitive option of optimization is to update $\mathcal{W}$ and $\alpha$ simultaneously on the training set, but the simultaneous optimization will cause $\alpha$ to overfit on the training set, which might derive the pruned result with poor generalization. Therefore, we leverage the bi-level optimization in the differentiable indicator search procedure, with $\alpha$ as the upper-level variable and $\mathcal{W}$ as the lower-level variable.
Specifically, it searches for $\alpha^*$ that minimizes the combination of validation loss $\mathcal{L}_{\text{val}}(\mathcal{W}^*(\alpha), \alpha)$ and regularizers $\mathcal{R}(\alpha)$, where the values of model parameters $\mathcal{W^*}$ are obtained by minimizing the training loss $\mathcal{L}_{\text{train}}(\mathcal{W}, \alpha)$:
\begin{align}
    \min_{\alpha} & \qquad \mathcal{L}_{\text{val}}(\mathcal{W}^*(\alpha), \alpha) + \lambda \mathcal{R}(\alpha), \\
    \textbf{s.t.} & \quad \mathcal{W}^*(\alpha) = \arg \min_{\mathcal{W}} \mathcal{L}_{\text{train}}(\mathcal{W}, \alpha).
\end{align}

To solve the bi-level optimization problem, $\mathcal{W}$ and $\alpha$ are updated in a multi-step scheme by gradient descent on training and validation sets, and finally reach the local minima. The alternating gradient updating process resembles~\cite{liu2019darts} with more details provided in the reference paper.

\subsection{Annealing-Relaxed Channel Indicator}\label{sec:method:aci}

To make the relaxed channel indicator a better approximation of binarized channel indicator, the value range for each entry $\widetilde{I}_l^i$ should be limited between 0 and 1.
Hence, a straightforward solution is to use a sigmoid function over auxiliary parameters, formally:
\begin{equation}\label{eqn:simple_ci}
    \widetilde{I}_l^i = \frac{1}{1 + e^{-\alpha_l^i}}.
\end{equation}

However, the above approximation has two inevitable problems.
(1) The resulting $\widetilde{I}_l^i$ might not converge to sparse values, i.e., close to 0 or 1, as there is no guarantee for sparsity based on the continuous auxiliary parameters.
\rebuttal{Hence, a hand-crafted threshold is still required to binarize the final pruning result, which limits the robustness of this method and violates the idea of automatic channel pruning.}
(2) Moreover, the discretization brings the non-negligible discrepancy between the continuous search result and binarized model~\cite{arber2019understanding,chu2019fair,liu2019darts}. This could deteriorate the accuracy of the pruned model when joint parameters of $W$ and $\alpha$ are stuck in sharp local minima where a small perturbation can lead to large performance degradation.

We propose an annealing-relaxed channel indicator to fill the discrepancy.
Concretely, we add a temperature variable $T$ on the sigmoid function, which is set to be high at the beginning and gradually anneals to zero at the end of training. As a result, the indicator is initially continuous to support the gradient update of the auxiliary parameters, and finally converges to the binarized state that leads to the pruned model: 
\begin{equation}
    \widetilde{I_l^i} = H_T(\alpha_l^i) = \frac{1}{1 + e^{-\alpha_l^i/T}}, \quad  I_l^i = \lim_{T \rightarrow 0}H_T(\alpha_l^i),
\end{equation}
where $\alpha_l^i$ represents the auxiliary parameter. 
The annealing-relaxed function $H_T(\cdot)$ is initialized with a high temperature $T = T_0$. 
\rebuttal{When the training proceeds into $n$-th epoch, the temperature of $H_T$ anneals to $T_0/\psi(n)$, where $\psi(\cdot)$ denotes the temperature annealing scheme.} 
Finally, the discrete $I_l^i$ can be approached with $T \rightarrow 0$ at the end of the search.

\subsection{Structural Restrictions by Regularizers}\label{sec:method:sr}

The vanilla cross-entropy loss itself is infeasible to induce \emph{a priori} structural restrictions, e.g., the number of floating point operations (FLOPs), which play a critical role in pruning. Therefore, we introduce three regularizers into the search procedure when updating the auxiliary parameters.

\dif{\para{Lasso regularizer.} 
A naturally used sparsity regularizer is the lasso regularizer:
\begin{equation}
    \mathcal{R}_{\text{lasso}}=\sum_{l=1}^{L}\sum_{i=1}^{c_l}\big|\big|H_T(\alpha_l^i)\big|\big|_{1}.
\end{equation} 

$\mathcal{R}_{\text{lasso}}$ uses the lasso/L1 regularization and can effectively zero out some channel indicators, but the pruning rate resulted from $\mathcal{R}_{\text{lasso}}$ is not controllable and heavily dependent on its weight on the whole loss function.}

\para{Continuous FLOPs estimator regularizer.} The value of the annealing-relaxed channel indicator $H_T(\alpha_l^{i})$ \rebuttal{can} be viewed as the probability of preserving the corresponding channel in the final pruned model. With the annealing-relaxed indicator of all output channels, we \rebuttal{can} estimate the expectation of overall FLOPs of the pruned model by accumulating FLOPs of each channel. 
In the $l$-th convolutional layer, the continuous FLOPs estimator could be represented as:
\dif{
\begin{equation}
    E_{\text{FLOPs}}(\alpha)= \sum_{l=1}^{L} \bigg(p_l \cdot \Big(\sum_{i=1}^{c_{l-1}}H_T\big(\alpha_{l-1}^i\big)\Big) \cdot \Big(\sum_{i=1}^{c_{l}}H_T\big(\alpha_l^i\big)\Big) \bigg)
\end{equation}
}
where $p_l=h_l \times w_l \times k_l^2$, and $k_l$ denotes the kernel size, $h_l$ and $w_l$ denote the spatial size of the output feature maps.

Inspired by \thirdrebuttal{Dong \textit{et al.}}~\cite{NIPS2019_8364}, we build up the regularizer given the continuous FLOPs estimator:
\begin{equation}
    \mathcal{R}_{\text{FLOPs}}(\alpha)=\left\{\begin{array}{cc}\log (E_{\text{FLOPs}}(\alpha)), &\frac{E_{\text{FLOPs}}(\alpha)}{F} > 1 \\ -\log (E_{\text{FLOPs}}(\alpha)), &\frac{E_{\text{FLOPs}}(\alpha)}{F} < 1 - \epsilon \\ 0, &otherwise \end{array}\right.
\end{equation}
where $F$ denotes the expected FLOPs and $\epsilon \ll 1$. The continuous FLOPs estimator regularizer guides DAIS to search for the optimized pruned model with FLOPs in range of $[(1-\epsilon)*F, F]$.

\dif{
\para{Symmetry regularizer.} 
Residual blocks are widely integrated in recent network designs, which greatly improve the capability of gradient propagation across multiple layers. The residual block is realized by adding a shortcut connection of non-parameterized identity mapping from input to output when the numbers of input and output channels are equal. 
However, the existing channel pruning methods~\cite{wu2019fbnet, NIPS2019_8364} cannot guarantee the equality between the numbers of input and output channels of a residual block, and they either drop the shortcut~\cite{wu2019fbnet} or replace the identity function with a $1 \times 1$ convolutional layer~\cite{NIPS2019_8364} for dimension matching.
These replacements might increase the difficulty of gradient propagation across multiple layers and lead to the problem of vanishing and exploding gradients~\cite{he2016deep}.

Therefore, we propose a symmetry regularizer for channel pruning on networks with residual connections, which keeps the consistency between the input and output channels within a residual block and reserves the identity mapping. The symmetry regularizer is defined as:
\begin{equation}
    \mathcal{R}_{\text{sym}} = \sum_{(l,l')}\bigg|\Big(\sum_{i=1}^{c_l}H_T(\alpha_l^i)\Big) - \Big(\sum_{i=1}^{c_{l'}}H_T(\alpha_{l'}^i)\Big)\bigg|, 
\end{equation}
where $(l, l')$ denotes a residual block with $c_l$ input channels and $c_{l'}$ output channels. We also explore the importance of the identity mapping on deep residual networks in Sec.~\ref{sec:exp:sr}.
}

\begin{table*}[t]
\centering
\caption{Comparison on channel pruning approaches using ResNet on the CIFAR-10 and CIFAR-100 datasets. ``Pruning Acc'' = accuracy of the pruned model, ``Acc Drop'' = accuracy drop, ``FLOPs'' = FLOPs (pruning ratio). We display two pruning result of DAIS for each model, where the first result is for best accuracy and the latter one is for best pruning rate.
}
\resizebox{0.95\textwidth}{!}{
\begin{tabular}{cc|ccc|ccc}
\toprule
\multirow{2}{*}{Depth}
& 
\multirow{2}{*}{Method}
& \multicolumn{3}{c|}{CIFAR-10} & \multicolumn{3}{c}{CIFAR-100} \\
& & Pruning Acc & Acc Drop & FLOPs & Pruning Acc & Acc Drop & FLOPs \\
\midrule

\multirow{7.5}{*}{32}
& LCCL~\cite{dong2017more}  & 90.74\% & 1.59\%  & 4.76E7 (31.2\%) & 67.39\% & 2.69\% & 4.32E7 (37.5\%) \\
& SFP~\cite{he2018soft}  & 92.08\% & 0.55\%  & 4.03E7 (41.5\%) & 68.37\% & 1.40\% & 4.03E7 (41.5\%) \\
& FPGM~\cite{he2019filter}  & 92.31\% & \thirdrebuttal{\textbf{0.32\%}}  & 4.03E7 (41.5\%) & 68.52\% & 1.25\% & 4.03E7 (41.5\%) \\
& CNN-FCF~\cite{li2019compressing} & 92.18\% & 1.07\% & 3.99E7 (42.2\%) & - & - & - \\
& TAS~\cite{NIPS2019_8364}  & 93.16\%	& 0.73\%  & 3.50E7 (49.4\%) & \thirdrebuttal{\textbf{72.41\%}} & \thirdrebuttal{\textbf{-1.80\%}} & 4.25E7 (38.5\%) \\
& LFPC~\cite{He_2020_CVPR}  & 92.12\%	& 0.51\%  & 3.27E7 (49.4\%) & - & - & - \\

\cmidrule(lr){2-8}
& DAIS & \thirdrebuttal{\textbf{93.49\%}}	& 0.57\%  & 3.19E7 (\textbf{53.9\%}) & 72.20\% & -1.04\% & 3.94E7 (\textbf{42.9\%}) \\

\midrule
\multirow{13.5}{*}{56}
& LCCL~\cite{dong2017more}  & 92.81\% & 1.54\%  & 7.81E7 (37.9\%) & 68.37\% & 2.96\% & 7.63E7 (39.3\%) \\
& AMC~\cite{he2018amc}  & 91.90\% & 0.90\%  & 6.29E7 (50.0\%) & - & - & -  \\
& SFP~\cite{he2018soft}  & 93.35\% & 0.56\%  & 5.94E7 (52.6\%) & 68.79\% & 2.61\% & 5.94E7 (52.6\%) \\
& FPGM~\cite{he2019filter}  & 93.49\% & 0.42\%  & 5.94E7 (52.6\%) & 69.66\% & 1.75\% & 5.94E7 (52.6\%) \\
& CNN-FCF~\cite{li2019compressing} & 93.38\% & \thirdrebuttal{\textbf{-0.24\%}} & 7.20E7 (42.8\%) & - & - & - \\
& TAS~\cite{NIPS2019_8364}  & \thirdrebuttal{\textbf{93.69\%}}	& 0.77\%  & 5.95E7 (52.7\%) & 72.25\% &  0.93\% & 6.12E7 (51.3\%) \\
& GBN~\cite{you2019gate}  & 93.07\% & 0.03\%  & 3.72E7 (70.3\%) & - & - & - \\
& GAL~\cite{lin2019towards}  & 93.38\% & 0.12\%  & 7.83E7 (37.6\%) & - & - & - \\
& LFPC~\cite{He_2020_CVPR}  & 93.24\% & 0.35\%  & 5.91E7 (52.9\%) & 70.83\% & \thirdrebuttal{\textbf{0.58\%}} & 6.08E7 (51.6\%) \\
& HRank~\cite{Lin_2020_CVPR}  & 93.17\% & 0.35\%  & 6.27E7 (50.0\%) & - & - & - \\
& ABCPruner~\cite{LinJZZW020} & 93.23\% & 0.03\% & 5.84E7 (54.1\%) & - & - & - \\
& SCP~\cite{icml2020_1485}  & 93.23\% & 0.46\%  & 6.10E7 (51.5\%) & - & - & - \\

\cmidrule(lr){2-8}

& DAIS & 93.53\%	& 1.00\%  & 3.64E7 (\textbf{70.9\%}) & \thirdrebuttal{\textbf{72.57\%}} &  0.81\% & 5.84E7 (\textbf{53.6\%}) \\

\midrule
\multirow{9.5}{*}{110}
& LCCL~\cite{dong2017more}  & 93.44\% & 0.19\%  & 1.68E8 (34.2\%) & 70.78\% & 2.01\% & 1.73E8 (31.3\%) \\
& SFP~\cite{he2018soft}  & 92.97\% & 0.70\%  & 1.21E8 (52.3\%) & 71.28\% & 2.86\% & 1.21E8 (52.3\%) \\
& FPGM~\cite{he2019filter}  & 93.85\% & -0.17\%  & 1.21E8 (52.3\%) & 72.55\% & 1.59\% & 1.21E8 (52.3\%) \\
& CNN-FCF~\cite{li2019compressing} & 93.67\% & -0.09\% & 1.44E8 (43.1\%) & - & - & - \\
& TAS~\cite{NIPS2019_8364}  & 94.33\% & 0.64\%  & 1.19E8 (53.0\%) & 73.16\% &  1.90\% & 1.20E8 (52.6\%) \\
& GAL~\cite{lin2019towards}  & 92.74\% & 0.76\%  & 1.30E8 (48.5\%) & - & - & - \\
& LFPC~\cite{He_2020_CVPR}  & 93.07\% & 0.61\%  & 1.01E8 (60.0\%) & - & - & - \\
& HRank~\cite{Lin_2020_CVPR}  & 93.36\% & 0.87\%  & 1.06E8 (58.2\%) & - &  - & - \\
\cmidrule(lr){2-8}
& DAIS & \thirdrebuttal{\textbf{95.02\%}} & \thirdrebuttal{\textbf{-0.60\%}}  & 1.01E8 ({\textbf{60.0\%}}) & \thirdrebuttal{\textbf{74.69\%}} & \thirdrebuttal{\textbf{-0.65\%}} & 1.14E8 (\textbf{56.7\%}) \\
\bottomrule
\end{tabular}
}
\label{tab:cifar}
\end{table*}

\begin{table}[t]
\centering
\caption{Results of light-weighted networks on CIFAR-10. ``M.Net'' and ``R-20'' denote the ResNet-20 and MobileNet respectively.}
\resizebox{\linewidth}{!}{
\dif{
\begin{tabular}{cc|c|c|c}
\toprule
& & Pruning Acc & Acc Drop & FLOPs \\
\midrule
\multirow{3.5}{*}{M.Net}
&M.Net-$0.75$ & 91.65\% & 1.22\% & 1.95E8 (43.3\%)\\
&CGNet~\cite{hua2019channel} & 87.56\% & \thirdrebuttal{\textbf{0.29\%}} & 1.19E8 (65.3\%) \\
\cmidrule(lr){2-5}
&DAIS & \thirdrebuttal{\textbf{91.87\%}} & 1.00\% &1.15E8 (\textbf{66.6\%})\\
\midrule
\multirow{6.5}{*}{R-20}
&LCCL~\cite{dong2017more}  & 91.68\% & 1.06\%  & 2.61E7 (36.0\%) \\
&SFP~\cite{he2018soft}  & 90.83\% & 1.37\%  & 2.43E7 (42.2\%) \\
&FPGM~\cite{he2019filter}  & 91.09\% & 1.11\%  & 2.43E7 (42.2\%) \\
&CNN-FCF~\cite{li2019compressing} & 91.13\% & 1.07\% & 2.38E7 (41.6\%) \\
&TAS~\cite{NIPS2019_8364}  & 92.88\% & 0.00\%  & 2.24E7 (45.0\%) \\
\cmidrule(lr){2-5}
&DAIS & \thirdrebuttal{\textbf{92.89\%}} & \thirdrebuttal{\textbf{-0.36\%}} & 1.91E7 (\textbf{51.1\%})  \\ 
\bottomrule
\end{tabular}
}
}
\label{tab:lightweighted}
\end{table}

\begin{table*}[t]
\centering
\caption{Comparison with baseline methods on ImageNet with ResNet-18/34/50. ``Latency'' = Latency (speedup).}
\resizebox{0.85\textwidth}{!}{
\begin{tabular}{cc|cc|cc|c|c}
\toprule
\multirow{2}{*}{Depth} & \multirow{2}{*}{Method}
&\multicolumn{2}{c|}{Top-1} & \multicolumn{2}{c|}{Top-5} & \multirow{2}{*}{FLOPs} & \multirow{2}{*}{Latency}\\

&&  Prune Acc &  Acc Drop &  Prune Acc &  Acc Drop &  & \\
\midrule
\multirow{5.5}{*}{18}
&Uniform & 66.12\% & 3.64\% & 87.25\% & 1.83\% & 1.06E9 (41.8\%) & 0.21s (1.52$\times$)\\
&LCCL~\cite{dong2017more}  & 66.33\% & 3.65\%  & 86.94\% & 2.29\% &  1.19E9 (34.6\%) & - \\
&SFP~\cite{he2018soft}  & 67.10\% & 3.18\% & 87.78\% & 1.85\% & 1.06E9 (41.8\%) & -\\
&ABCPruner~\cite{LinJZZW020} & 67.28\% & 2.38\% & - & - & 1.01E9 (\thirdrebuttal{\textbf{44.9\%}}) & - \\
\cmidrule(lr){2-8}
&DAIS & \textbf{67.56\%} & \thirdrebuttal{\textbf{2.20\%}} & \textbf{87.90\%} & \thirdrebuttal{\textbf{1.18\%}} & 1.03E9 (43.3\%) & 0.19s (\thirdrebuttal{\textbf{1.68$\times$}})\\
\midrule
\multirow{6.5}{*}{34}
&Uniform & 71.38\% & 1.93\% & 90.52\% & 0.90\% & 2.13E9 (41.9\%) & 0.33s (1.83$\times$)\\
&PFEC~\cite{li2016pruning}  & 72.17\% & 1.06\% & - & - & 2.78E9 (24.2\%) & -\\
&SFP~\cite{he2018soft}  & 71.83\% & 2.09\% & 90.33\% & 1.29\% & 2.16E9 (41.1\%) & -\\
&FPGM~\cite{he2019filter}  & 72.54\% & 1.38\%  & \thirdrebuttal{\textbf{91.13\%}} & 0.49\% & 2.16E9 (41.1\%) & -\\
&ABCPruner~\cite{LinJZZW020} & 70.98\% & 2.30\% & - & - & 2.17E9 (41.0\%) & - \\
\cmidrule(lr){2-8}
&DAIS & \textbf{72.77\%} & \thirdrebuttal{\textbf{0.54\%}} & 90.99\% & \thirdrebuttal{\textbf{0.43\%}} & 2.13E9 (\textbf{41.9\%}) & 0.31s (\thirdrebuttal{\textbf{1.93$\times$}})\\
\midrule
\multirow{6.5}{*}{50}
&SFP~\cite{he2018soft}  & 62.14\% & 14.0\% & 84.60\% & 8.27\% & 2.38E9 (41.8\%) & -\\
&SSS~\cite{huang2018data}  & 71.82\% & 4.30\% & 90.79\% & 2.07\% & 2.33E9 (43.0\%) & -\\
&FPGM~\cite{he2019filter}  & 74.13\% & 2.02\%  & 91.94\% & 0.93\% & 1.90E9 (53.5\%) & -\\
&ABCPruner~\cite{LinJZZW020} & 73.86\% & 2.15\% & 91.69\% & 1.27\% & 1.89E9 (54.1\%) & - \\
& SCP~\cite{icml2020_1485} & 74.20\% & \thirdrebuttal{\textbf{1.69\%}} & 92.00\% & 0.98\% & 1.87E9 (54.3\%) & -\\
\cmidrule(lr){2-8}
&DAIS & \textbf{74.45\%} & 1.70\% & \textbf{92.21\%} & \thirdrebuttal{\textbf{0.66\%}} & 1.83E9 (\textbf{55.3\%}) & 0.31s (\thirdrebuttal{\textbf{1.77}$\times$})\\
\bottomrule
\end{tabular}
}
\label{tab:imagenet}
\end{table*}

\begin{table}[t]
\centering
\caption{Comparison with baseline methods on ImageNet with VGG-16. The pruning results of $2\times, 4\times$ speedup ratio are reported.}
\resizebox{\linewidth}{!}{
\dif{
\begin{tabular}{c|cc|cc}
\toprule
\multirow{3}{*}{Method}
&\multicolumn{2}{c|}{$2\times$} & \multicolumn{2}{c}{$4\times$} \\
& \tabincell{c}{Top-1\\Acc Drop} & \tabincell{c}{Top-5\\Acc Drop} & \tabincell{c}{Top-1\\Acc Drop} & \tabincell{c}{Top-5\\Acc Drop} \\
\midrule
Jaderberg~\textit{et al.}~\cite{JaderbergVZ14} & - & - & - & 9.70\% \\
Zhang~\textit{et al.}~\cite{ZhangZHS16} & - & - & - & 3.84\%  \\
Li~\textit{et al.}~\cite{li2016pruning} & - & - & - & 8.60\%  \\
SSS~\cite{huang2018data} & - & - & 3.93\% & 2.64\% \\
CC-GAP~\cite{Li_2021_CVPR} & 2.78\% & 1.68\% & - & - \\
\cmidrule(lr){1-5}
DAIS & \thirdrebuttal{\textbf{1.64\%}} & \thirdrebuttal{\textbf{0.79\%}} & \thirdrebuttal{\textbf{2.91\%}} & \thirdrebuttal{\textbf{1.48\%}} \\
\bottomrule
\end{tabular}
}
}
\label{tab:imagenet:vgg}
\end{table}

\section{Experiments}
\subsection{Experiment Setup}
\para{Datasets.}
We evaluated the effectiveness of DAIS on CIFAR-10~\cite{krizhevsky2009learning}, CIFAR-100~\cite{krizhevsky2009learning}, and ImageNet ILSVRC-12~\cite{deng2009imagenet}. CIFAR-10 and CIFAR-100 both consist of 50K training images and 10K test images, and they have 10 and 100 classes, respectively.
ImageNet ILSVRC-12 contains 1280K training images and 50K test images for 1000 classes. 

\begin{figure}[t]
    \centering
    \includegraphics[width=\linewidth]{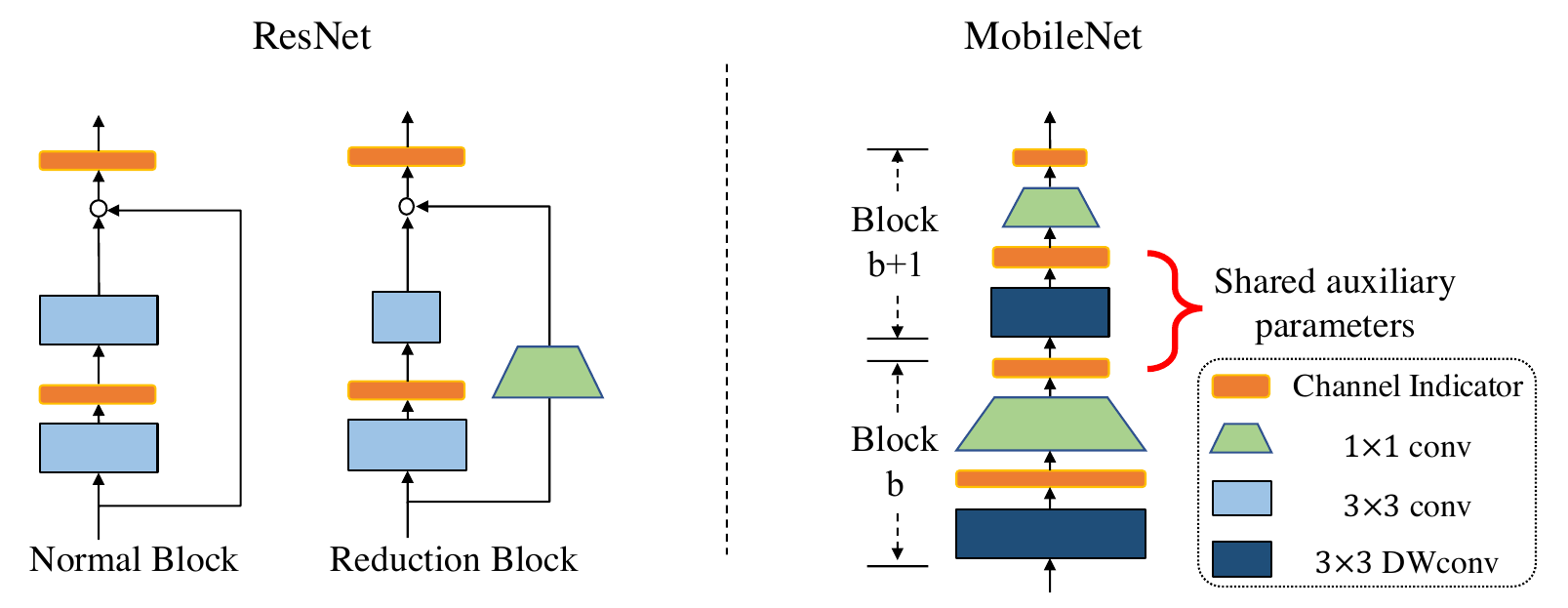}
    \caption{The channel indicator setup for ResNet and MobileNet networks.}
    \label{appendix:fig:channel_indicator}
\end{figure}

\para{Channel indicator setting.} We implemented the annealing-relaxed channel indicator to each convolutional layer except for the first layer, conforming to baselines~\cite{dong2017more}. The implementation details of the annealing-relaxed indicator are as follows. On ResNet, there are two kinds of residual blocks: normal block and reduction block. Most residual blocks in the network are normal blocks, and the reduction block is employed to down-sample the spatial size of the feature map while doubling the number of the output channel. We adopted the same implementation for these two types of blocks: the first annealing channel indicator is used right behind the first $3\times3$ convolution, and the second indicator is implemented after the addition of two paths. On MobileNet, we built up the annealing-relaxed channel indicator for each $1\times1$ convolution and $3\times3$ depth-wise convolution (DWconv). As the DWconv requires the same number of input and output channels, the annealing-relaxed indicator is shared before and after the DWconv. The implementation details are shown in Fig.~\ref{appendix:fig:channel_indicator}.

\para{Differentiable annealing indicator search.} In the search process, we used $\mathcal{R}_{\text{FLOPs}}$ and $\mathcal{R}_{\text{sym}}$ as the default regularizers. For the update of the auxiliary parameters, DAIS adopted Adam~\cite{KingmaB2015adam} as the optimizer with the momentum of $(0.5, 0.999)$, where the learning rate and weight decay rate were both set to 1E-3. For the update of the network parameters, DAIS leveraged the SGD optimizer with $0.9$ momentum, and the learning rate was initialized with $0.1$ and reduced by the cosine scheduler~\cite{loshchilov2016sgdr}. On the CIFAR experiments, we implemented the differentiable annealing indicator search for 100 epochs. The original training images were divided into the training set and validation set with the $7:3$ ratio. On the ImageNet experiments, we searched 7508 iterations with batch size 256.

For all experiments, the auxiliary parameters were initialized with a normal distribution $\alpha \in \mathcal{N}(1, 0.1^2)$. The initialization temperature was $T_0=1$, and the temperature annealing scheme was $\psi(n)=49\times n/N_{\text{max}}+1$, where $N_{\text{max}}$ denotes the total number of training epochs. The batch size of the search procedure was 256. The weights of $\mathcal{R}_{\text{FLOPs}}$ was $2$ and $\epsilon=0.05$. The weights of $\mathcal{R}_{\text{sym}}$ was $0.01$ for ResNet-56/110 and $0$ otherwise. The weight decay of the SGD optimizer on the CIFAR and ImageNet experiments were 5E-5 and 1E-5 respectively. 

\para{Fine-tuning.}
For all experiments, we used SGD as the optimizer with a momentum of $0.9$. The batch size was 256, and the learning rate was initialized with $0.1$. For the CIFAR experiments, we trained \thirdrebuttal{each model for 300 epochs}. The learning rate was warmed up with 5 epochs and then reduced with the cosine scheduler. We adopted random crop, random horizontal flipping and \thirdrebuttal{random erasing~\cite{Zhong0KL020}} as default augmentations. We trained all models with a single NVIDIA P40 GPU in the CIFAR experiments. For the ImageNet experiments, we trained 120 epochs for each model, and the learning rate was decayed by 10 every 30 epochs. We used the random resized crop and random horizontal flipping for data augmentation. Besides, conforming to the methods~\cite{he2019filter, NIPS2019_8364} that use original models to implicitly-or-explicitly transfer the knowledge to pruned models, we used the feature distillation method~\cite{Heo_2019_ICCV} on the ImageNet experiments. We leveraged 4 NVIDIA P40 GPUs in the ImageNet experiments.

\subsection{Comparisons with State-of-the-Art Methods}

\para{Results on CIFAR.}
We first evaluated the performance of DAIS on ResNet-32/56/110 with CIFAR-10 and CIFAR-100. As illustrated in TABLE~\ref{tab:cifar}, DAIS consistently outperformed various state-of-the-art pruning approaches on CIFAR. For ResNet-32 on CIFAR-100, DAIS reduced 42.9\% FLOPs and increased accuracy by 1.04\% compared with the original network. Besides, the differentiable annealing indicator search was efficient: DAIS only spent 95 minutes on the search procedure, while TAS finished the search in 228 minutes. For ResNet-56 on CIFAR-10 and CIFAR-100, DAIS obtained fewer FLOPs than other pruning methods. Furthermore, since DAIS was a one-shot solution, it required less training effort than the iterative pruning approaches like GBN~\cite{you2019gate}. In the case of deeper architecture ResNet-110, on both CIFAR-10 and CIFAR-100, our DAIS obtained the best FLOPs reduction ratio, accuracy drop, and accuracy, indicating that the proposed symmetry regularizer perfectly improved the capability of gradient propagation in the deeper layers.

We also evaluated the performance of DAIS on light-weighted networks like MobileNet~\cite{howard2017mobilenets} and ResNet-20. For MobileNet, DAIS got better accuracy with fewer FLOPs compared with MobileNet-$0.75$ and CGNet~\cite{hua2019channel}, as shown in TABLE~\ref{tab:lightweighted}. For ResNet-20 on CIFAR-10, DAIS reduced 51.1\% FLOPs and increased accuracy by 0.36\% compared with the original network. It also outperformed state-of-the-art methods like SFP~\cite{he2018soft} and FPGM~\cite{he2019filter}, which revealed that the simple intuitive pruning rates design (evenly pruning for each layer) could not surpass the automatic search-based pruning approach in DAIS. 

\dif{
\para{Results on ImageNet.}
To evaluate the effectiveness of DAIS, extensive experiments were performed on ResNet-18/34/50~\cite{he2016deep} and VGG-16 with the ImageNet dataset. \thirdrebuttal{TABLE}~\ref{tab:imagenet} reports the accuracy, pruning rate and latency of the pruned models on ResNet-18/34/50. The latency is calculated by the implementation on Galaxy S9 with PyTorch Mobile\footnote{\url{https://pytorch.org/mobile/home/}}. For ResNet-18, the derived model got $1.68 \times$ speedup on \rebuttal{Galaxy} S9, and it exceeded the uniform pruning model by 1.44\% accuracy with fewer FLOPs. For ResNet-34, the searched model by DAIS got minimized accuracy drop and maximized pruning ratio compared with baseline methods. On deep models like ResNet-50, DAIS still outperformed other methods. The pruning results of VGG-16 are shown in \thirdrebuttal{TABLE}~\ref{tab:imagenet:vgg}. DAIS got less Top-1/5 accuracy drop compared with other methods on $2\times, 4\times$ speedup ratio of FLOPs. These experimental results verified the generality of DAIS on large-scale datasets. 
}

\dif{
\subsection{Importance of Identity Mapping on Deep Residual Network}\label{sec:exp:sr}

To verify the importance of identity mapping in the shortcut connection, we built up two network design spaces~\cite{radosavovic2020designing}, where multiple network instances were randomly generated with 110 layers, simulating channel pruning results of ResNet-110. All these network instances were trained 300 epochs on the CIFAR-10 dataset. Specifically, the ``Random'' design space had no constraint on the number of output channels, while in ``Constrained'' design space, every residual block was restricted to have the same number of input and output channels. The details of the two network design spaces are as follows:

\begin{figure}[t]
  \centering
    \includegraphics[width=0.6\columnwidth]{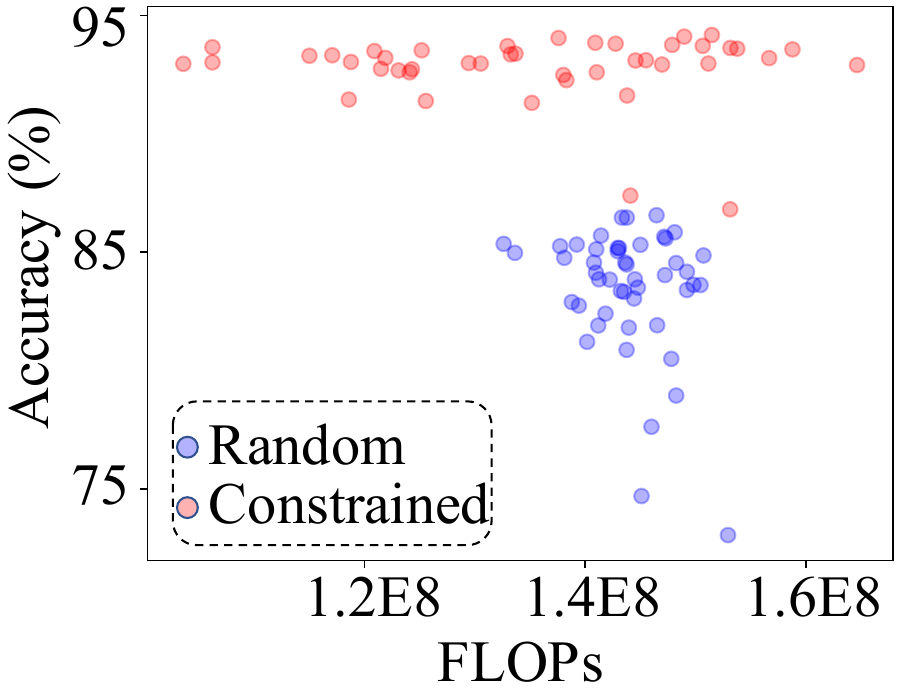}
  \caption{Statistics of 100 network instances generated from the ``Random'' and ``Constrained'' design spaces.}
  \label{fig:sym}
\end{figure}

\begin{itemize}
    \item \textbf{``Random''}. For each pruned model instance,  we randomly sampled the number of output channels $c_l^{\text{Random}}=c_l*\mathcal{U}(0.5, 1)$, where $c_l$ denotes the number of output channels in the $l$-th layer of the original ResNet-110, and $\mathcal{U}(0.5, 1)$ denotes a uniform sampling function ranging from 0.5 to 1.
    \item \textbf{``Constrained''}. For the model instances in the ``Constrained'' design space, we hope to reserve most of the identity functions in the shortcut path of the residual block, which requires the same number of input and output channels in each residual block. Therefore, we designed the sampling procedure as follows. For all residual blocks in the first stage (output feature map size: $32 \times 32$), we set the output channels to be $16*\mathcal{U}(0.5, 1)$. For all residual blocks in the second stage (output feature map size: $16 \times 16$), we set the output channels to be $32*\mathcal{U}(0.5, 1)$. For all residual blocks in the third stage (output feature map size: $8 \times 8$), we set the output channels to be $64*\mathcal{U}(0.5, 1)$. The rest convolutional layers were randomly sampled \textit{in the same pattern} as ``Random'': $c_l^{\text{Constrained}}=c_l*\mathcal{U}(0.5, 1)$.
\end{itemize} 

As shown in Fig.~\ref{fig:sym}, With similar FLOPs $F \in [1.1\text{E}8, 1.5\text{E}8]$, the model instances generated from the ``Constrained'' design space obtained a consistent performance improvement (in terms of accuracy and pruning ratio) compared with the ``Random'' design space, indicating the importance of consistency between the input and output channels within a residual block.
}

\subsection{Ablation Study}\label{sec:exp:ab}

\para{Comparison with other search methods.} 
To verify the effectiveness of the search procedure of DAIS, we compared the differentiable search with three other search methods. 
``Slimming'' denoted the method~\cite{liu2017learning} in which the channel indicator was represented by the weights of BN layers, and the channels with small BN weights would be filtered out.
``Random'' and ``Constrained'' are the methods mentioned in Sec.~\ref{sec:exp:sr}, where 50 model instances were randomly generated by each method and the instances with the best accuracy were collected in TABLE~\ref{tab:ab}. DAIS outperformed these three methods on both the accuracy and the pruning rate, verifying the effectiveness of the differentiable annealing indicator search.

\begin{table}[t]
\centering
\caption{Results of ablation study with ResNet-110 on CIFAR-10. ``FLOPs'' = FLOPs (pruning ratio).}
\resizebox{0.98\linewidth}{!}{
\begin{tabular}{c|c|c|c}
\toprule
& Pruning Acc & Acc Drop & FLOPs \\
\midrule
Slimming & 84.94\% & 9.48\% & 1.15E8 (54.46\%)\\
Random & 86.56\% & 7.86\% & 1.46E8 (42.15\%)\\
Constrained & 94.18\% & 0.24\% & 1.51E8 (40.17\%)\\
DAIS & \textbf{95.02\%} & \thirdrebuttal{\textbf{-0.60\%}} & 1.01E8 (\textbf{60.00\%}) \\
\midrule
w/o $\mathcal{R}_{\text{sym}}$ & 88.97\% & 5.45\% & 1.20E8 (52.62\%)\\
w/o $\mathcal{R}_{\text{FLOPs}}$ & 89.15\% & 5.27\% & 1.45E8 (42.92\%) \\
DAIS & \textbf{95.02\%} & \thirdrebuttal{\textbf{-0.60\%}} & 1.01E8 (\textbf{60.00\%}) \\
\midrule
w/o annealing & 93.57\% & 0.85\% & 1.54E8 (39.17\%) \\
DAIS & \textbf{95.02\%} & \thirdrebuttal{\textbf{-0.60\%}} & 1.01E8 (\textbf{60.00\%}) \\
\midrule
w/o bi-level & 94.73\% & -0.31\% & 1.04E8 (59.07\%) \\
DAIS & \textbf{95.02\%} & \thirdrebuttal{\textbf{-0.60\%}} & 1.01E8 (\textbf{60.00\%}) \\
\bottomrule
\end{tabular}
}
\label{tab:ab}
\end{table}

\para{Effectiveness of $\mathcal{R}_{\text{FLOPs}}$ and $\mathcal{R}_{\text{sym}}$.}
We conducted additional experiments to verify the advantages of $\mathcal{R}_{\text{FLOPs}}$ and $\mathcal{R}_{\text{sym}}$. In TABLE~\ref{tab:ab}, ``w/o $\mathcal{R}_{\text{FLOPs}}$'' denotes a model variant that replaced $\mathcal{R}_{\text{FLOPs}}$ by  $\mathcal{R}_{\text{lasso}}$. The replacement led to a huge accuracy drop with less pruning rate, which indicated the effectiveness of $\mathcal{R}_{\text{FLOPs}}$. The second variant ``w/o $\mathcal{R}_{\text{sym}}$'' removed the symmetry regularizer from DAIS and suffered a $6.05\%$ accuracy reduction. The comparison revealed the importance of the symmetry regularizer on the ultra-deep residual networks. 

\begin{figure}[t]
  \centering
    \includegraphics[width=0.95\linewidth]{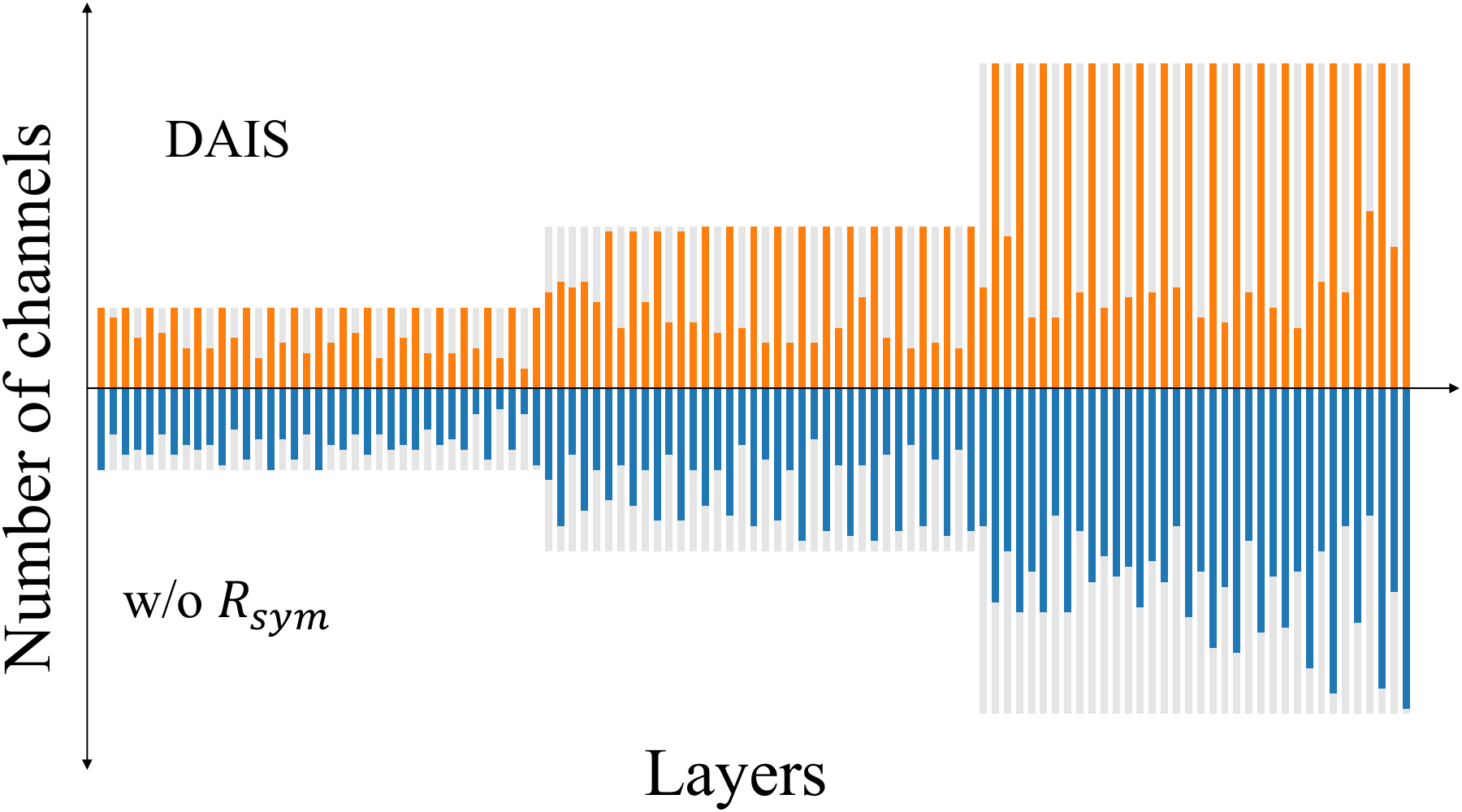}
  \caption{The pruned model for ResNet-110 on CIFAR-10.}
  \label{fig:resnet110}
\end{figure}

Furthermore, we visualized the pruning result of DAIS and ``w/o $\mathcal{R}_{\text{sym}}$'' in Fig.~\ref{fig:resnet110}. Compared to the pruned model derived from ``w/o $\mathcal{R}_{\text{sym}}$'', in each residual block, the pruned model derived from DAIS was intended to prune more channels in the first $3\times3$ convolution, and reserved most channels in the second $3\times3$ convolution. Therefore, the pruned model could reserve most identity functions in the shortcut path, which improved the capability of gradient propagation across multiple layers. 

\para{Effectiveness of the annealing-relaxed function.}
With the help of the annealing-relaxed function, the channel indicators converge to the binarization automatically with no hand-crafted threshold. In this experiment, we explored the influences of removing the annealing-relaxed function. We designed a variant ``w/o annealing'' which kept the channel indicator in Eq.~(\ref{eqn:simple_ci}) fixed without any temperature annealing. Without temperature annealing, the $\widetilde{I}_l^i$ could not converge to the binarization automatically, and therefore we manually filtered out the channels with $\widetilde{I}_l^i < 0.55$. Results showed that DAIS performed better in terms of both accuracy and pruning ratios, verifying the necessity of the annealing-relaxed function.

\para{Effectiveness of the bi-level optimization.}
We implemented the bi-level optimization on the differentiable annealing indicator search, which updated the model parameters on the training set and updated the channel indicator on the validation set. To verify the necessity of the bi-level optimization, we designed the variant ``w/o bi-level'' in TABLE~\ref{tab:ab}, which jointly optimized the model parameters and channel indicators on the training set. The pruned model searched by ``w/o bi-level'' might overfit the training data, and results showed that it performed worse on both accuracy and pruning ratio than DAIS, verifying the effectiveness of the bi-level optimization.

\begin{table*}[t]
\centering
\caption{The indicator search procedure on ResNet-20 on CIFAR-10 dataset. The number of output channels with $\widetilde{I}_l^i>0.5$ are collected. Some channels will be firstly pruned and then recovered in the search procedure.}
\resizebox{\textwidth}{!}{
\begin{tabular}{c|ccccccccccccccccccc}
\toprule
Epochs&\multicolumn{19}{c}{Number of Output Channels with $\widetilde{I}_l^i>0.5$ (From Layer 1 to Layer 19). } \\
\midrule
0&16&16&16&16&16&16&16&32&32&32&32&32&32&64&64&64&64&64&64\\
20&16&3&15&2&16&0&16&26&32&13&32&12&32&61&64&61&62&46&64 \\
40&16&3&15&4&16&2&15&26&30&17&30&14&32&53&64&54&60&42&58 \\
60&16&3&14&4&16&5&16&25&30&17&29&14&31&52&64&51&57&39&57 \\
80&16&4&14&5&16&6&15&24&31&17&29&16&31&52&63&48&56&37&57 \\
100&16&7&14&4&16&7&15&24&30&17&29&13&31&52&63&47&56&34&57 \\
\bottomrule
\end{tabular}
}
\label{appendix:tab:recoverable}
\end{table*}

\subsection{Case Study}
\para{Recoverability.}
According to \thirdrebuttal{Guo \textit{et al.}}~\cite{guo2016dynamic}, over-pruning or incorrect pruning might happen in the pruning process and lead to degraded performance, and therefore the recoverability is necessary for a good pruning algorithm. Therefore, we visualized the differentiable annealing indicator search procedure on ResNet-20 in TABLE~\ref{appendix:tab:recoverable}. For every 20 epochs, we collected the number of output channels with $\widetilde{I}_l^i>0.5$. We observed that the number of output channels did not decrease monotonically, and some channels were recovered as the training proceeded. For example, in the second convolutional layer, the number of output channels was decreased to 3, and then recovered to 7. The recoverability revealed that DAIS could rectify the mistakes of over-pruning or incorrect pruning, which might also explain the effectiveness of DAIS.

\dif{
\para{Robustness of DAIS.}
We first explored the impact of a shorter training scheme. The ``DAIS.e50'' trained 50 epochs for the differentiable annealing indicator search procedure and got similar performance with the original DAIS, indicating the efficiency of DAIS. The second experiment explored the impact of different temperature decay schemes. ``DAIS.cosine'' leveraged a consine decay scheme $\psi(n) = 49\times (1-cos(\frac{\pi}{2}n/N_{\text{max}})) + 1$, while ``DAIS.smallT'' adopted a smaller termination temperature by $\psi(n)=99\times n/N_{\text{max}} + 1$. All the corresponding results were shown in TABLE~\ref{tab:robustness}. We also explored the effect of the $\alpha$ initialization in TABLE~\ref{tab:init}. All variants got similar performance, which verified the robustness of DAIS to the normal temperature decay schemes and $\alpha$ initialization.
}

\begin{table}[t]
\centering
\caption{Comparisons on training schemes and temperature decay schemes with ResNet-110 on CIFAR-10. ``FLOPs'' = FLOPs (pruning ratio).}
\resizebox{0.98\linewidth}{!}{
\begin{tabular}{c|c|c|c}
\toprule
& Pruning Acc & Acc Drop & FLOPs \\
\midrule
DAIS.e50 & 94.40\% & 0.02\% &  1.02E8 (59.69\%) \\
\cmidrule(lr){1-4}
DAIS.cosine & 95.00\% & -0.58\% & 1.04E8 (59.06\%) \\
DAIS.smallT & 94.61\% & -0.19\% & 1.05E8 (58.70\%) \\
\cmidrule(lr){1-4}
DAIS & \textbf{95.02\%} & \thirdrebuttal{\textbf{-0.60\%}} & \thirdrebuttal{1.01E8} (\textbf{60.00\%}) \\
\bottomrule
\end{tabular}
}
\label{tab:robustness}
\end{table}

\begin{table}[t]
\centering
\caption{Comparisons on different $\alpha$ initialization with ResNet-56 on CIFAR-10. ``FLOPs'' = FLOPs (pruning ratio).}
\resizebox{0.98\linewidth}{!}{
\dif{
\begin{tabular}{c|c|c|c}
\toprule
& Pruning Acc & Acc Drop & FLOPs \\
\midrule
DAIS.$\mathcal{N}(0, 0.1^2)$ & 93.11\% & 1.42\% & 5.41E7 (57.01\%)\\
DAIS.$\mathcal{N}(0, 0.05^2)$ & 93.36\% & 1.17\% & 5.34E7 (\thirdrebuttal{\textbf{57.54\%}})\\
DAIS.$\mathcal{N}(1, 0.05^2)$ & 92.97\% & 1.56\% & 5.55E7 (55.89\%) \\
\cmidrule(lr){1-4}
DAIS & {\textbf{93.71\%}} & \thirdrebuttal{\textbf{0.82\%}} & 5.61E7 (\thirdrebuttal{55.40\%}) \\
\bottomrule
\end{tabular}
}
}
\label{tab:init}
\end{table}

\para{One-shot capability of DAIS.}
We conducted an iterative pruning scheme on MobileNet, denoted by ``Iterative-$i$'' in \rebuttal{Table~\ref{tab:iterative}}. Similar to most iterative pruning schemes~\cite{liuautocompress,zhu2017prune}, the pruning rate was set to be large (45\%) in the first round and gradually increased 10\% in latter rounds. The experimental result showed that the original one-shot DAIS obtained slightly better performance than  ``Iterative-$3$'', indicating the differentiable search procedure could effectively search the pruned models in a one-shot manner. 

\dif{
\para{Trade-off between accuracy and FLOPs.}
DAIS can perform fine-grained search on the pruned model with a target FLOPs $F$ by $\mathcal{R}_{\text{FLOPs}}$, which controls the trade-off between FLOPs and accuracy.
We derived models with different pruning rates on ResNet-56 formatted by (pruning rate, accuracy): (27.43\%,94.91\%), (48.85\%,94.24\%), (55.36\%, 93.71\%), (70.92\%, 93.53\%).
}

\begin{table}[t]
\centering
\caption{Results of iterative pruning of DAIS on CIFAR-10. ``Iterative-$i$'' denotes the $i$-th round pruning result by DAIS.}
\resizebox{\linewidth}{!}{
\begin{tabular}{c|c|c|c}
\toprule
& Pruning Acc & Acc Drop & FLOPs \\
\midrule
MobileNet & 92.87\% & $-$ & 3.44E8 \\
\cmidrule{1-4}
Iterative-$1$ & 91.66\% &1.21\% & 1.88E8 (45.34\%) \\
Iterative-$2$ & 91.76\% &1.11\% & 1.50E8 (56.40\%)\\
Iterative-$3$ & 91.77\% & 1.10\% & 1.18E8 (65.70\%)\\
\cmidrule(lr){1-4}
DAIS & \textbf{91.87\%} & \thirdrebuttal{\textbf{1.00\%}} &1.15E8 (\textbf{66.60\%})\\
\bottomrule
\end{tabular}
}
\label{tab:iterative}
\end{table}

\subsection{Transfer Learning}
We have demonstrated the effectiveness of DAIS in classification tasks, and we explored its performance on some other computer vision tasks like semantic segmentation and scene text recognition. 

\para{Semantic segmentation.} We selected DeepLabV3+~\cite{chen2018encoder} with ResNet-34 as the baseline model and performed experiments on PASCAL VOC 2012~\cite{pascal-voc-2012}. As suggested by \thirdrebuttal{Liu \textit{et al.}}~\cite{liu2018rethinking}, we first pruned the ResNet-34 backbone with DAIS on ImageNet, and then finetuned a DeepLabV3+ with the pruned ResNet-34 as its backbone on PASCAL VOC 2012. In the finetuning process on PASCAL VOC 2012 dataset, we followed the same training strategies as the original DeepLabV3+ paper. All models were trained for 50 epochs with the training set of PASCAL VOC 2012 and an augmentation dataset~\cite{hariharan2011semantic}. The performance was measured in terms of Pixel Accuracy (PA), Mean Pixel Accuracy (MPA), Mean Interaction over Union (MIoU) and Frequency Weighted Interaction over Union (FWIoU). As illustrated in TABLE~\ref{tab:segmentation}, DAIS reduced 15.39\% FLOPs on DeepLabV3+ \thirdrebuttal{with small pixel accuracy drop} ($92.87\% \rightarrow 92.51\%$), indicating that the pruned model derived by DAIS could be transferred into another task without major performance changes. 

\begin{table}[t]
\centering
\caption{Semantic segmentation results based on DeepLabV3+ on PASCAL VOC 2012.}
\label{tab:segmentation}
\resizebox{\linewidth}{!}{
\begin{tabular}{c|c|c|c|c|c}
\toprule
& PA & MPA & MIoU & FWIoU & FLOPs \\
\midrule
DeepLabV3+ & 92.87\% & 81.71\% & 72.52\% & 87.16\% & 5.67E10 \\
\cmidrule(lr){1-6}
Uniform & 91.39\% & 73.63\% & 68.42\% & 84.22\% & 4.94E10 (12.78\%) \\
DAIS & \thirdrebuttal{\textbf{92.51\%}} & \thirdrebuttal{\textbf{79.23\%}} & \thirdrebuttal{\textbf{71.29\%}} & \thirdrebuttal{\textbf{86.48\%}} & 4.80E10 (\thirdrebuttal{\textbf{15.39\%}}) \\
\bottomrule
\end{tabular}
}
\end{table}

\dif{
\para{Scene text recognition.} We also explored the performance of DAIS on scene text recognition tasks. We adopted STN~\cite{jaderberg2015spatial,long2019rethinking} with ResNet-34 as the baseline model. Similar to the segmentation experiments in Sec. 4.5, we replaced the original ResNet-34 with the pruned model searched by DAIS on ImageNet~\cite{deng2009imagenet}. All models were trained on SynthText~\cite{Gupta16} and Synth90K~\cite{Jaderberg14c} and evaluated on 7 real-world datasets: IIIT5k~\cite{mishra2012scene}, SVT~\cite{wang2011end}, IC03~\cite{lucas2003icdar}, IC13~\cite{karatzas2013icdar},  IC15~\cite{karatzas2015icdar}, SVTP~\cite{quy2013recognizing} and CUTE~\cite{risnumawan2014robust}. We adopted the same training hyperparameters as Aster~\cite{shi2018aster}. DAIS had the pruning rate of 17.06\% and the uniform pruning method prunes 16.98\% FLOPs. As illustrated in TABLE~\ref{tab:ocr}, DAIS outperformed the uniform pruning on 4 datasets with less FLOPs. These experiments indicated the generality of our method on several vision tasks including image classification, semantic segmentation and scene text recognition.
}

\begin{table}[t]
\centering
\caption{Scene text recognition accuracies based on STN across 7 real world datasets.}
\resizebox{\linewidth}{!}{
\begin{tabular}{c|ccccccc}
\toprule
& IIIT5k & SVT & IC03 & IC13 & IC15 & SVTP & CUTE \\
\midrule
STN & 92.9\% & 88.4\% & 91.9\% & 89.5\% & 78.3\% & 78.9\% & 78.5\% \\
\cmidrule(lr){1-8}
Uniform & \thirdrebuttal{\textbf{93.4\%}} & 88.9\% & 91.7\% & \thirdrebuttal{\textbf{89.3\%}} & \thirdrebuttal{\textbf{79.0\%}} & 78.6\% & 76.4\% \\
DAIS & 93.2\% & \textbf{89.8\%} & \textbf{93.2\%} & \thirdrebuttal{\textbf{89.3\%}} & 78.8\% & \textbf{78.8\%} & \textbf{79.2\%} \\
\bottomrule
\end{tabular}
}
\label{tab:ocr}
\end{table}

\section{Discussions}
 DAIS focuses on the automatic channel pruning, which also implicitly decreases the latency, memory footprint, and energy consumption of the original models. Specifically, \rebuttal{DAIS reduces the FLOPs, model parameters, and the memory footprint.
} As shown in TABLE~\ref{tab:imagenet}, the pruned models also have lower latency on mobile devices. In terms of energy consumption, DAIS has a shorter search time compared to other methods~\cite{he2018amc,NIPS2019_8364}, which implies lower CO2 marginal emission costs and cloud computing costs according to APQ~\cite{WangWCLL0LH20}. Besides, DAIS has lower calculation cost and calculation delay as shown in \thirdrebuttal{TABLE}~\ref{tab:imagenet}, which also decrease the CO2 emissions following \cite{abs-2104-08702}.

\section{Conclusion}
In this paper, we propose Differentiable Annealing Indicator Search (DAIS), which leverages the channel indicator to represent the sparsity and searches for an appropriate pruned model with the computation cost constraints. Specifically, DAIS approximates the binarized channel indicator with the annealing-relaxed indicator and then jointly optimizes the indicator and model parameters with gradient-based bi-level optimization. The annealing-relaxed indicator will automatically converge to the binarized state as the optimization proceeds and temperature anneals. Furthermore, DAIS proposes a continuous FLOPs estimator regularizer to precisely constrain model sizes and a symmetry regularizer to optimize the gradient propagation on very deep residual networks. Experimental results show that DAIS outperforms state-of-the-art methods on CIFAR-10, CIFAR-100, and ImageNet on different architectures, verifying the effectiveness of the differentiable search.

\section*{Acknowledgment}
This work is partially supported by National Key R\&D Program of China (2020YFB2103801), Beijing Academy of Artificial Intelligence (BAAI), NSFC (National Natural Science Foundation of China) 62032003, and BJNSF (Beijing Municipal Natural Science Foundation) L192004.

{\small
\bibliographystyle{IEEEtran}
\bibliography{Refs}
}

%

\begin{IEEEbiography}[{\includegraphics[width=1in,height=1.25in,clip,keepaspectratio]{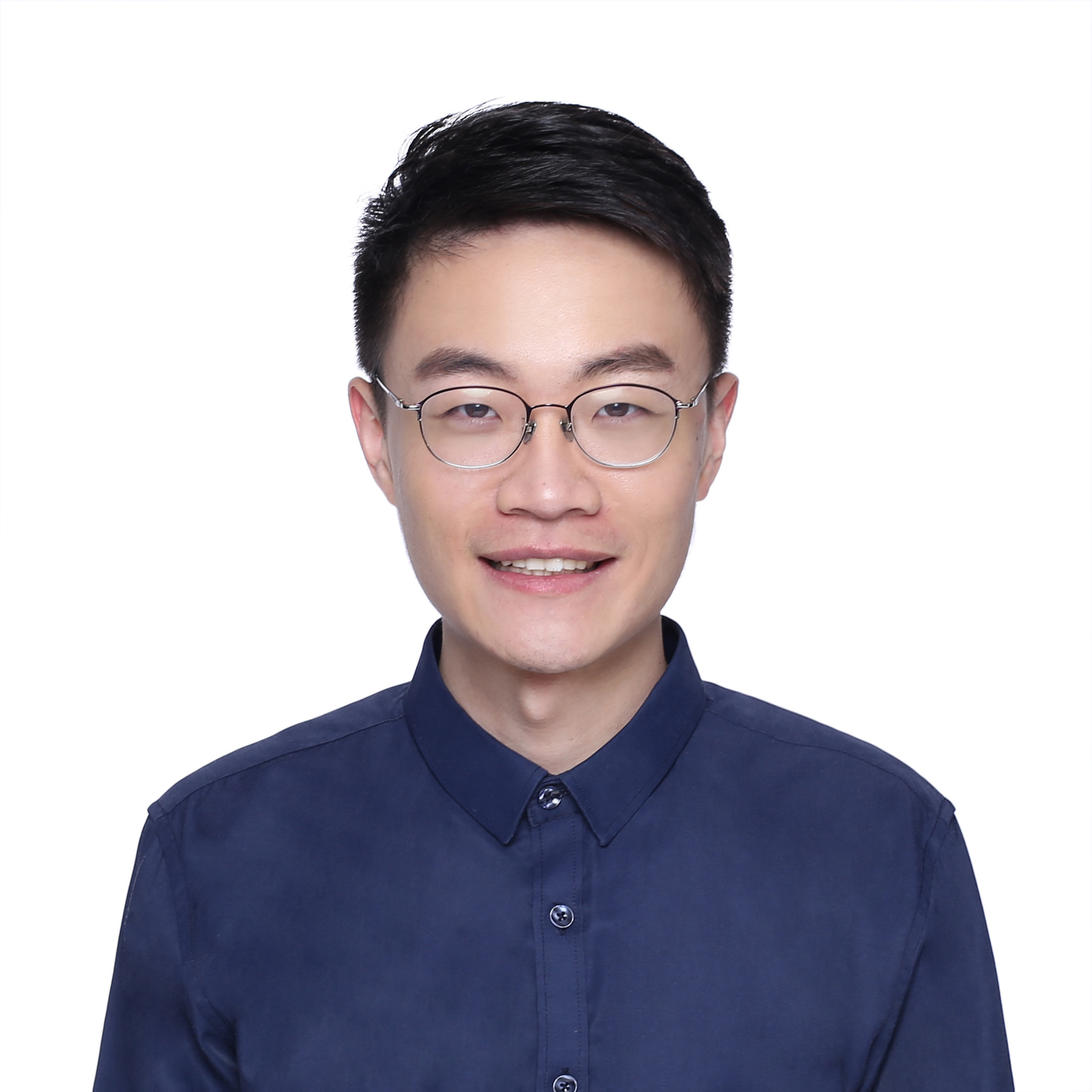}}]{Yushuo Guan} received the Master degree at the School of Electronics Engineering and Computer Science in Peking University, China, in 2021. He received his Bachelor degree from Peking University in 2018. His research focuses on model compression, scene text recognition and other machine learning applications.
\end{IEEEbiography}

\begin{IEEEbiography}[{\includegraphics[width=1in,height=1.25in,clip,keepaspectratio]{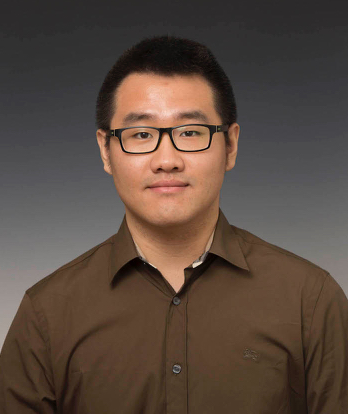}}]{Ning Liu} received the Ph.D. degree in computer engineering from the Northeastern University, Boston, MA, USA, in 2019. He is a researcher at Midea Group. His current research interests lie in deep learning, deep model compression and acceleration, deep reinforcement learning, and edge computing.
\end{IEEEbiography} 

\begin{IEEEbiography}[{\includegraphics[width=1in,height=1.25in,clip,keepaspectratio]{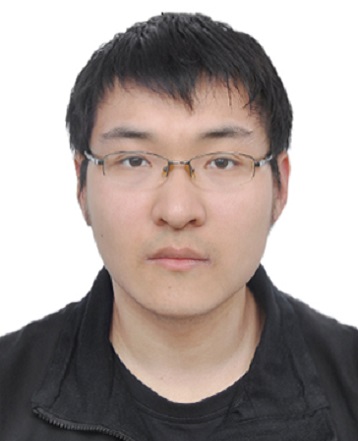}}]{Pengyu Zhao} received his Bachelor and Master degrees, both at the School of Electronics Engineering and Computer Science from Peking University, China, in 2017 and 2020 respectively. His research focuses on model compression, neural architecture search and other machine learning applications.
\end{IEEEbiography}

\begin{IEEEbiography}[{\includegraphics[width=1in,height=1.25in,clip,keepaspectratio]{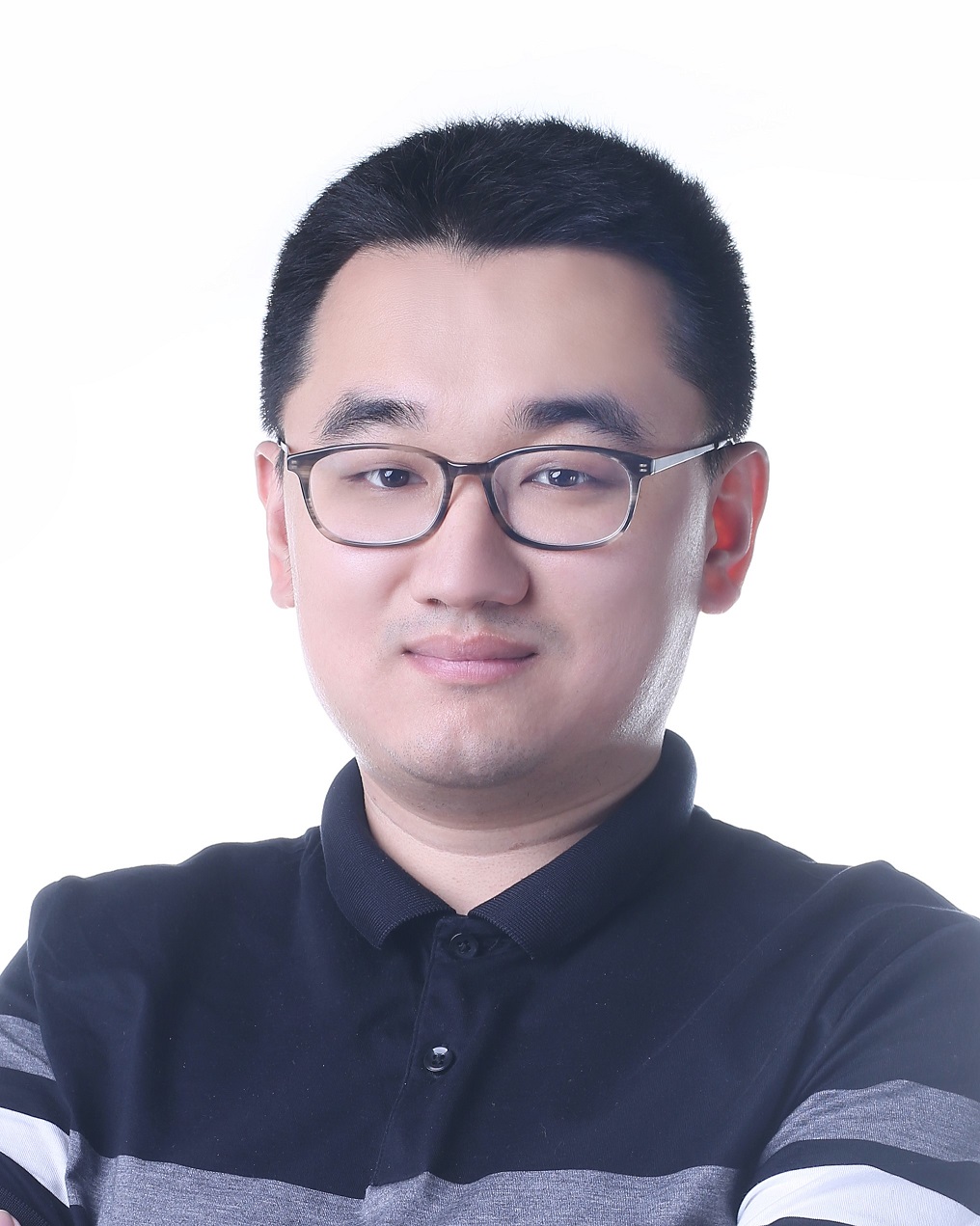}}]{Zhengping Che}
received the Ph.D. degree in Computer Science from the University of Southern California, Los Angeles, CA, USA, in 2018, and the B.Eng. degree in Computer Science from Tsinghua University, Beijing, China, in 2013.
He is now with AI Innovation Center, Midea Group.
His current research interests lie in the areas of machine learning, deep learning, computer vision, and time series analysis with applications to robot learning.
\end{IEEEbiography}

\begin{IEEEbiography}[{\includegraphics[width=1in,height=1.25in,clip,keepaspectratio]{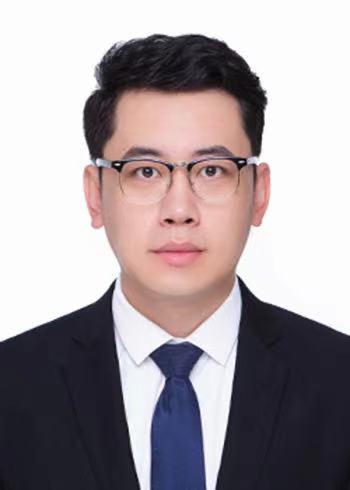}}]{Kaigui Bian}
received the PhD degree in computer engineering from Virginia Tech, Blacksburg, USA in 2011. He is currently an associate professor in the Institute of Network Computing and Information Systems, School of EECS, Peking University. His research interests include mobile computing, cognitive radio networks, network security, and privacy.
\end{IEEEbiography}

\begin{IEEEbiography}[{\includegraphics[width=1in,height=1.25in,clip,keepaspectratio]{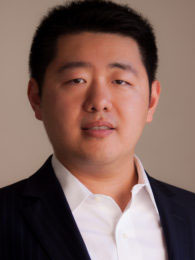}}]{Yanzhi Wang}
received the B.S. degree in electronic engineering from Tsinghua University, Beijing, China, in 2009, and the Ph.D. degree in computer engineering from the University of Southern California, Los Angeles, CA, USA, in 2014. His research interests include energy-efficient and high-performance implementations of deep learning and artificial intelligence systems, emerging deep learning algorithms/systems, generative adversarial networks, and deep reinforcement learning.

\end{IEEEbiography}

\begin{IEEEbiography}[{\includegraphics[width=1in,height=1.25in,clip,keepaspectratio]{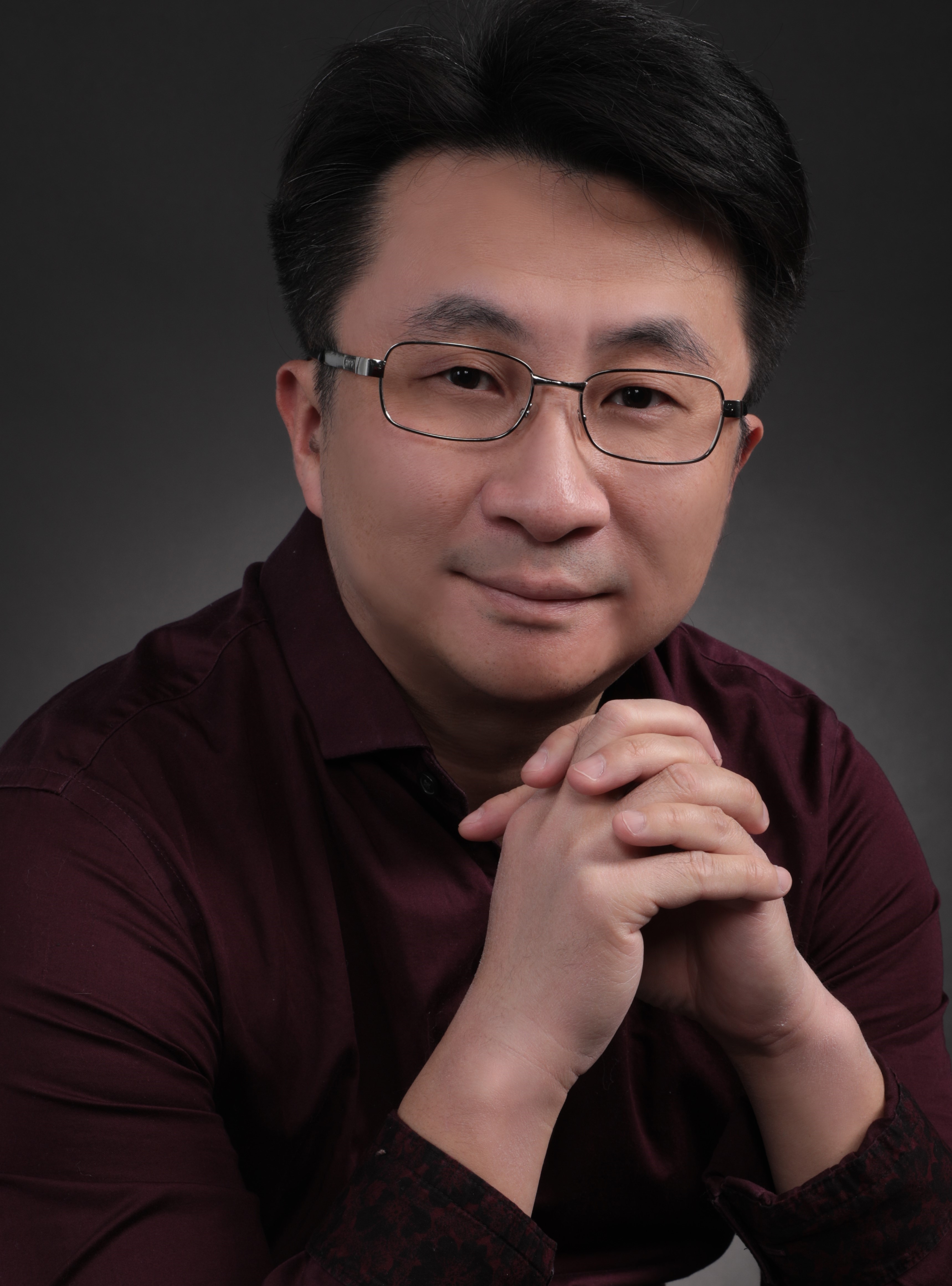}}]{Jian Tang}
received his Ph.D degree in Computer Science from Arizona State University in 2006. He is an IEEE Fellow and an ACM Distinguished Member. He is with Midea Group. His research interests lie in the areas of AI, IoT, Wireless Networking, Mobile Computing and Big Data Systems. Dr. Tang has published over 160 papers in premier journals and conferences. He received an NSF CAREER award in 2009. He also received several best paper awards, including the 2019 William R. Bennett Prize and the 2019 TCBD (Technical Committee on Big Data) Best Journal Paper Award from IEEE Communications Society (ComSoc), the 2016 Best Vehicular Electronics Paper Award from IEEE Vehicular Technology Society (VTS), and Best Paper Awards from the 2014 IEEE International Conference on Communications (ICC) and the 2015 IEEE Global Communications Conference (Globecom) respectively. He has served as an editor for several IEEE journals, including IEEE Transactions on Big Data, IEEE Transactions on Mobile Computing, etc. In addition, he served as a TPC co-chair for a few international conferences, including the IEEE/ACM IWQoS’2019, MobiQuitous’2018, IEEE iThings’2015. etc.; as the TPC vice chair for the INFOCOM’2019; and as an area TPC chair for INFOCOM 2017-2018. He is also an IEEE VTS Distinguished Lecturer, and the Chair of the Communications Switching and Routing Committee of IEEE ComSoc.
\end{IEEEbiography}




\end{document}